\documentclass[11pt,letterpaper]{article}

\usepackage{tabularx,booktabs}
\usepackage{url}
\usepackage{fullpage}
\usepackage{graphicx}
\usepackage{lipsum}
\newcommand{\junk}[1]{}

\usepackage{algorithm}
\usepackage{algpseudocode}
\usepackage{amsmath}
\usepackage{xcolor}
\usepackage{xcolor,colortbl}
\usepackage{todonotes}
\usepackage{amsfonts}
\usepackage[hidelinks]{hyperref}

\begin{document}
\thispagestyle{empty}

\title{CoViT: Real-time phylogenetics for the SARS-CoV-2 pandemic using Vision Transformers}
\author
{
\centering
Zuher Jahshan$^{1}$\footnote{Corresponding Author. Email: zuherjahshan@campus.technion.ac.il} 
\and Can Alkan$^{2}$
\and Leonid Yavits$^{3}$ \and
\\
		$^{1}$EE Dept, Technion - Israel Institute of Technology, Haifa, Israel\\
	   	$^{2}$Department of Computer Engineering, Bilkent University, Ankara, Turkey\\	
	   	$^{3}$EnICS Labs, Engineering Dept, Bar Ilan University, Ramat Gan, Israel\\
}

\date{}

\maketitle
\begin{abstract}
  Real-time viral genome detection, taxonomic classification and phylogenetic analysis are critical for efficient tracking and control of viral pandemics such as Covid-19. However, the unprecedented and still growing amounts of viral genome data create a computational bottleneck, which effectively prevents the real-time pandemic tracking. For genomic tracing to work effectively, each new viral genome sequence must be placed in its pangenomic context. Re-inferring the full phylogeny of SARS-CoV-2, with datasets containing millions of samples, is prohibitively slow even using powerful computational resources. We are attempting to alleviate the computational bottleneck by modifying and applying Vision Transformer, a recently developed neural network model for image recognition, to taxonomic classification and placement of viral genomes, such as SARS-CoV-2. Our solution, CoViT, places SARS-CoV-2 genome accessions onto SARS-CoV-2 phylogenetic tree with the accuracy of 94.2\%. Since CoViT is a classification neural network, it provides more than one likely placement. Specifically, one of the two most likely placements suggested by CoViT is correct with the probability of 97.9\%. The probability of the correct placement to be found among the five most likely placements generated by CoViT is 99.8\%. 
%CoViT reaches the top-1 accuracy of 97.1\% when placing a new SARS-CoV-2 genome in a context of 107 existing lineages, and 92.1\% when the number of lineage classes is 375.
The placement time is 0.055s per individual genome running on NVIDIA’s GeForce RTX 2080 Ti GPU. 
We make CoViT available to research community through GitHub: \url{https://github.com/zuherJahshan/covit}.
\end{abstract}

\section{Introduction}
  The COVID-19 pandemic exposed scientific infrastructures to an unprecedented global stress test. PCR-based testing alone required massive rearrangements in scientific communities and administrations. However, PCR-based testing only establishes a poor man’s solution to analyzing a pandemic: neither can it reveal emerging strains and lineages, nor can it accurately assign infections to phylogenetic clades. A successful pandemic monitoring and surveillance is only possible through sequencing and subsequent analysis of large amounts of viral genomes.

For the analysis to be efficient, each new viral genome must be analyzed in its pangenomic context, considering the entire evolutionary history of the virus. DNA sequencing progress made possible the creation of enormous SARS-CoV-2 genomic databases, with over 12 million sequenced SARS-CoV-2 genomes~\cite{gisaid} spread over thousands of distinct lineages. Evolutionary tracking can be accomplished by inferring
the full phylogeny for each such accession, but with  SARS-CoV-2 datasets containing millions of genomes, this becomes a computationally daunting challenge~\cite{turakhia2021ultrafast}. Alternatively, accessions can be placed in their pangenomic context by positioning them on an existing phylogenetic tree. 

Several recently developed tools aim at resolving this problem. EPA-ng~\cite{barbera2019epa} computes the optimal insertion position for an accession in a given reference phylogenetic tree with respect to its maximum-likelihood. IQ-TREE~\cite{minh2020iq} uses stochastic algorithms for estimating maximum likelihood phylogenies. UShER~\cite{turakhia2021ultrafast} is one of the latest phylogenetic placement tools. It uses a maximum parsimony approach where it searches for a placement that requires the fewest additional mutations.
%UShER is capable of classifying a SARS-CoV-2 genome into the correct sister node with an accuracy of 97.2\%~\cite{turakhia2021ultrafast}. 
UShER originally reported ~0.5s for a placement of a single accession onto the SARS-CoV-2 phylogenetic tree. However, our own experiment with the latest UShER revision 0.5.6 running on a different hardware (Intel i7-9700K 8 core machine), yielded a better placement time of 0.195s. An apparent limitation of UShER is that it only supports mutations caused by substitutions and it does not support mutations caused by indels.

In this work, we make a case for a different approach of placing viral genome accessions on an existing phylogenetic tree. We introduce and propose CoViT, an algorithm based on Vision Transformer (ViT)~\cite{dosovitskiy2020image}. ViT is a deep neural network for image recognition, which we modify and amend to enable extremely fast and accurate taxonomic classification of virus samples, and their placement onto the existing phylogeny. 
%We show that CoViT can quickly and accurately place new genomes onto a SARS-CoV-2 phylogeny.
%CoViT places newly acquired (query) samples onto the tree of SARS-CoV-2 lineages. 
Specifically, CoViT receives a SARS-CoV-2 genome and generates a list of most probable lineages ordered by their likelihood. One of the two most probable lineages identified by CoViT is the correct one with the probability of 97.9\%. Such probability grows to 99.8\% for five most probable lineages (i.e. the probability of the correct result being among these five most probable lineages is 99.8\%). The placement time is 0.055s on NVIDIA’s GeForce RTX 2080 Ti GPU, which is 3.53 times faster than the latest UShER revision. The placement time of CoViT is still 2.06 times faster than UShER (revision 0.5.6) when running on the same computational resources (i.e., Intel i7-9700K CPU with 8 cores, operating at 3.60GHz
with 32GB of DDR4 RAM).

We make CoViT available to research community through GitHub: 
\url{https://github.com/zuherJahshan/covit}, as well as at
\url{https://cov-lineages.org/resources.html}.

\section{Methods}
  \subsection{Background}
CoViT is based on ViT~\cite{dosovitskiy2020image}, a neural network primarily developed for image classification. We develop and apply a preprocessing step that employs MinHash~\cite{broder2000min}, whose role is extracting the informative feature vectors from the genome, which are further fed into a modified ViT for classification. In this section we provide a detailed introduction to the ViT and MinHash.

\subsubsection{Transformers}
\label{transformers}
\begin{figure*}[!t]
  \begin{center}
    \includegraphics[scale=0.3]{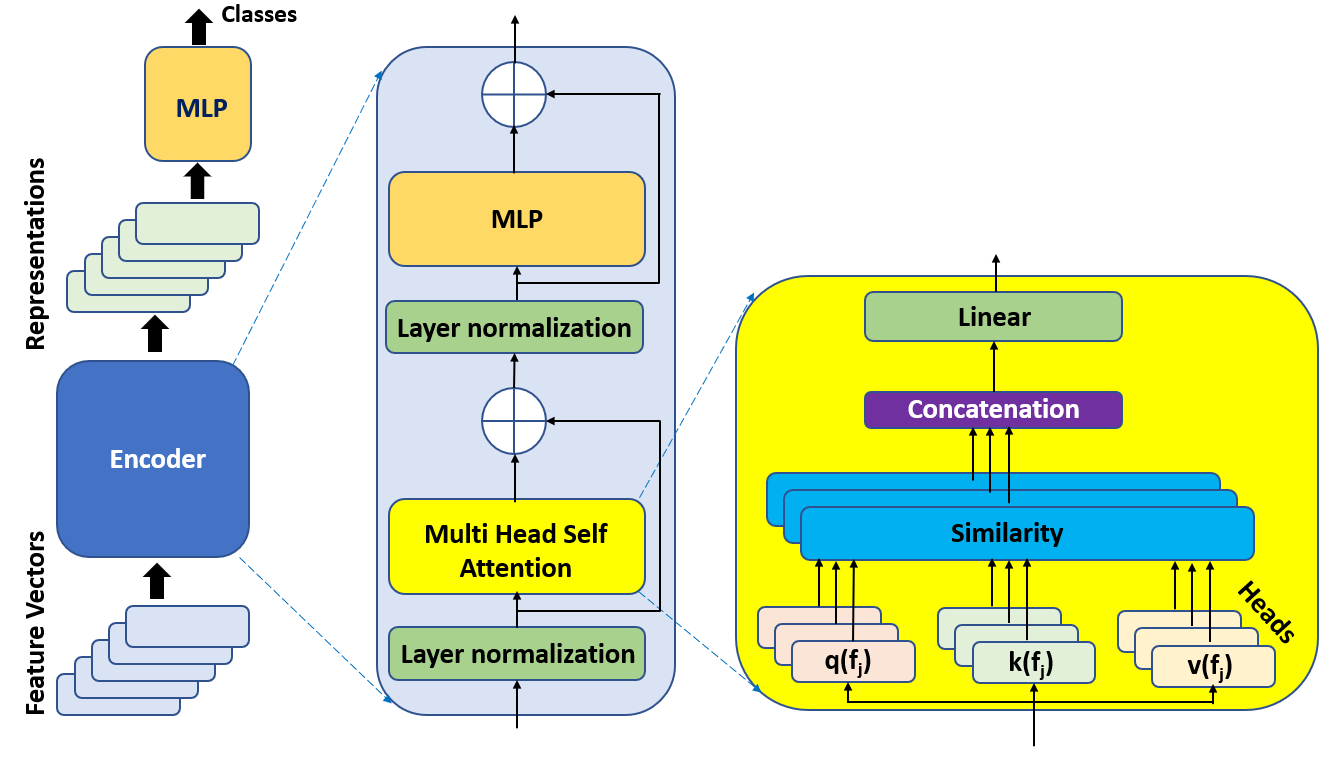}
  \end{center}
\caption{The Transformer encoder architecture is built as a stack of alternating $L$ multi-head self-attention (MHSA) layers and $L$ MLP blocks. Layer normalization is applied before every block and a residual connection is employed around each of the sub-layers (i.e., MHSA and MLP)}
\label{fig:TIC}
\end{figure*}

Transformer is a deep neural network model proposed by Vaswani et al.~\cite{vaswani2017attention}. It became state of the art in many deep learning applications, such as  Natural Language Processing. More recently, the ViT which is an image classification deep neural network based on the transformer model, was introduced in ~\cite{dosovitskiy2020image}. ViT architecture is presented in Figure~\ref{fig:TIC}. It comprises an encoder and a multilayer perceptron (MLP) layer with a softmax activation function to perform predictions (in this paper we will refer to this MLP layer as MLP head). An encoder receives a sequence of representations, each of them is of dimensionality $d_{model}$, and outputs a sequence of learnt representations of the same dimensionality. For the sake of clarity we refer to the sequence of input representations as ``feature vectors sequence'' throughout this section. The transformer encoder layer is composed of two main sub-layers, namely the Multi Head Self Attention (MHSA) and the piece-wise MLP. In the following subsections we will present the structure and the purpose of those sub-layers.

\textbf{Multi Head Self Attention:}
This sub-layer, depicted in Figure~\ref{fig:TIC} (right) receives as an input a sequence of feature vectors $f_1, f_2,\dots, f_n$, each of dimensionality $d_{model}$. The sub-layer outputs a sequence of learnt representations $r_1, r_2, \dots, r_n$, a representation per each feature vector. The output learnt representations are of the same dimensionality as the input ones.
The following presents the MHSA data flow. First, each feature vector $f_j$ is passed through $3h$ linear layers $\{q_l, k_l, v_l\}_{l=1}^{h}$, where $h$ denotes the number of heads, and 3 denotes the three types of different representations (i.e., query, key, and value). The purpose of those linear layers is to learn $h$ query representations for $f_j$, $h$ key representations for $f_j$ and $h$ value representations of $f_j$. Each linear layer defines a weight matrix as follows:
$W_l^q \in \mathbb{R}^{d_{model}\times d_{q}}$;\textbf{  } $ W_l^k \in \mathbb{R}^{d_{model}\times d_{k}}$;\textbf{  } $W_l^v \in \mathbb{R}^{d_{model}\times d_{v}}$.
They are used to calculate their respective representations:
$q_l(f_j) = f_j^T\cdot W_l^q$;\textbf{  } $k_l(f_j) = f_j^T\cdot W_l^k$;\textbf{  } $v_l(f_j) = f_j^T\cdot W_l^v$.

The query and key representations are used to calculate the degree of attention that the learnt output representation, $r_j$, will give to other feature vectors. The value representations are used to represent learnt properties of the feature vector that we intend to pass through the network. Formally, this first step of the MHSA layer is defined as follows:
\begin{align*}
    & A^l_j = \sum_{i=1}^{n}{similarity(q_l(f_j),k_l(f_i))}\cdot v_l(f_i)\\
    & similarity(q_l(f_j),k_l(f_i)) =
    softmax_i(\frac{q_l(f_j)^T k_l(f_1)}{\sqrt{d_k}}, \dots, \frac{q_l(f_j)^T k_l(f_1)}{\sqrt{d_k}})
\end{align*}
where $l$ refers to the $l$'th head, and $softmax_i$ is the $i$'th element of the softmax function, defined as
$softmax(x_1, x_2, \dots, x_n)=\frac{1}{\sum_{i=1}^{n}{e^{x_i}}}(e^{x_1}, e^{x_2},\dots, e^{x_n}).
$

Note that for a given head $l$, feature vectors with higher similarity to $f_j$ will have a larger dot product with $q_l(f_j)$ compared to other feature vectors. This drives the network to learn, for each representation, $r_j$, on which feature vectors of the input sequence it should focus its attention to learn the representations of value. To recap, representations $A_j^1, A_j^2 \dots, A_j^h$ depend on feature vectors that receive a high degree of attention. 

Next, these representations are concatenated to a form $A_j = A_j^1 \circ A_j^2 \circ \dots \circ A_j^h$. 

The last step of the MHSA is applying to $A_j$ a linear layer, to generate the MHSA output representation. 
%This output representation, as stated earlier is of dimensionality $d_{model}$.

\textbf{Trainable weights and hyper-parameters:}
For each head, $l\in [1, h]$, we define three weight matrices $W_l^q, W_l^k, W_l^v$.
Additional weight matrix $W^o\in \mathbb{R}^{h\cdot d_v\times d_{model}}$ is reserved for the linear layer that processes $A_j$. 
%It is easy to convince yourself that $W^o\in \mathbb{R}^{h\cdot d_v\times d_{model}}$.

The MHSA hyper-parameters include the number of heads $h$, the dimensionality of the query and key representation $d_q=d_k$, and the dimensionality of the value representation $d_v$.

\textbf{Piece-wise Multi Layer Perceptron:}
The piece-wise MLP receives as an input a sequence of representations (feature vectors). For each feature vector it learns an output representation. Unlike the MHSA, in the piece-wise MLP layer each learnt representation depends only on its input feature vector and does not depend on other feature vectors in the sequence. This layer transforms each input feature vector into an intermediate representation of dimensionality $d_{ff}$, and learns the output representation which is of dimensionality $d_{model}$. Formally, to perform this calculation, we define two weight matrices
$
W^{ff}\in \mathbb{R}^{d_{model}\times d_{ff}}$ and \textbf{  } $W^m\in \mathbb{R}^{d_{ff}\times d_{model}}$.
and calculate the representation of $f$ as follows:
\begin{align*}
    MLP(f) = (ReLU(f^T\cdot W^{ff}))^T\cdot W^m
\end{align*}
\textbf{Trainable weights and hyper-parameters:}
There are two weight matrices $W^{ff}$ and $W^m$ as defined above. The only hyper-parameter is $d_{ff}$.

\textbf{Layer Normalization:}
Layer normalization~\cite{ba2016layer} is a method used to normalize the activities of the neurons of a layer improving the training speed for neural network models. It directly estimates the normalization statistics from the summed inputs to the neurons within a hidden layer.

\textbf{The Transformer Encoder:}
The Transformer Encoder (TE) comprises a stack of L identical encoder layers, each comprising an MHSA sub-layer followed by a piece-wise MLP sub-layer. Layer normalization is applied before each sub-layer, and a residual connection is employed around each of the two sub-layers.

\textbf{Hyper-parameters of the TE:}
 The number of encoder layers - $L$, the dimensionality of the query and key representations - $d_q = q_k$, the dimensionality of the value representation - $d_v$, and the dimensionality of the intermediate representation of the piece-wise MLP - $d_{ff}$.

%%%%%%%%%%%%%%%%%%%%%%%%%%%%%%%%%%%%%%%%%%%%%%%%%%%%
%%%%%%%%%%%%%%%%%%% MinHash %%%%%%%%%%%%%%%%%%%%%%%%
%%%%%%%%%%%%%%%%%%%%%%%%%%%%%%%%%%%%%%%%%%%%%%%%%%%%
\subsubsection{The MinHash Scheme}
The min-wise independent permutations (MinHash)~\cite{broder2000min} is a technique for similarity estimation. 
%It was initially developed to detect duplicate web pages~\cite{broder2000identifying}. More recently, the 
MinHash was applied in many tasks in computational biology, including genome assembly~\cite{berlin2015assembling, shafin2020nanopore}, 
%gene clustering~\cite{yang2011parallel, drew2014strand}, 
metagenomic gene clustering~\cite{rasheed2013map, muller2017metacache} and genomic distance estimation~\cite{ondov2016mash}. MinHash implements the following sketch function:

Given a set of characters $A$, a compression factor $n$, and a hash function $h$, the elements of the set $A$ are hashed using function $h$ to generate the set $H(A)$. Then the elements of $H(A)$ are sorted, and the smallest $n$ elements are returned as described in Algorithm~\ref{alg:sketch}.

\begin{algorithm}
\caption{The sketch algorithm $sketch(A)$}\label{alg:sketch}
\begin{algorithmic}
\Require set $A = \{a_1,\dots,a_{|A|}\}$, compression parameter $n$ and a hash function $h$
\State $sketch \gets \emptyset$
\State $H(A) \gets (h(a_1), \dots h(a_{|A|}))$
\State Sort H(A) to get $(h(a_{i_1}), h(a_{i_2}), \dots, h(a_{i_{|A|}}))$\\
\Return $(a_{i_1}, a_{i_2}, \dots, a_{i_{n}})$\footnotemark
\end{algorithmic}
\end{algorithm}
\footnotetext{These are the genome representative kmers referred to in Section 2.2.1}

%The sketch algorithm, effectively returns a subset of $A$, designated $sketch(A)$. 
The subset obtained by applying the sketch algorithm provides a good estimate for the Jaccard index~\cite{hancock2004jaccard}, defined as follows: Given two sets $A, B$ the Jaccard index is 
$
J(A, B) = \frac{|A\cap B|}{|A\cup B|}
$.

Formally, given two sets, A and B, $\frac{|sketch(A) \cap sketch(B)|}{|sketch(A) \cup sketch(B)|} \approx J(A, B)$. 
We use the sketch algorithm to extract feature vectors from a genome, as presented further.

\subsection{CoViT Architecture}

\begin{figure}[!t]
  \begin{center}
    \includegraphics[scale=0.48]{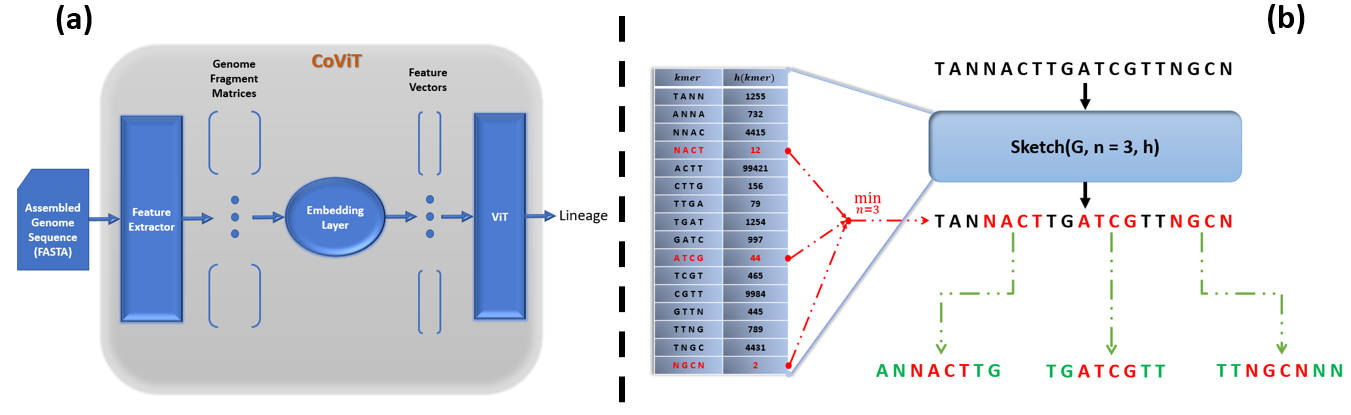}
  \end{center}
\caption{(a): A general overview of the CoViT Architecture. %The genome sequence is first passed through the feature extractor which generates genome fragment matrices. Then each matrix passes through an embedding layer that transforms the one-hot encoding to numerical tokens. Lastly the embedded fragments are fed to the vision transformer and it returns the placement. 
(b): Feature Extractor workflow example. 
%$4-mers$ are extracted from the genome sequence using MinHash, where we apply $sketch$ and keep $n=3$ minimal hashed $4-mers$. Afterwards those $4-mers$ are used as anchors to generate the fragments (which are the 4-mers extended by 2 basepairs in each direction). The last step is one-hot encoding of the fragments.
%probabilities of the query genome being found in each lineage class
}
\label{fig:covitarch}
\label{fig:fee}
\end{figure}

The top-level architecture of CoViT is shown in Figure~\ref{fig:covitarch}(a). The input for CoViT is an assembled genome in FASTA file format. To convert the genome into a sequence of feature vectors, we build a pipeline of preprocessing steps, comprising the feature extractor and the embedding layer. The feature extractor chooses a set of fragments (i.e., genome subsequences of fixed length), represents each fragment as a two-dimensional matrix and outputs those genome fragment matrices. Each genome fragment matrix is fed into an embedding layer consisting one neuron. This layer converts the genome fragment matrix into a genome fragment vector, designated the feature vector. The feature vectors are fed into the ViT which outputs the most likely placement candidates (by attaching probabilities to each lineage).

The feature extractor together with the embedding layer transform a genome into a sequence of representative feature vectors, which are the numerical representations of genome fragments. We implement the feature extractor using MinHash to preserve similarities in genome. It allows the feature extractor to find features in the query genome that are shared by the known lineages. The embedding layer is used as a dimensionality reduction step. It allows to reduce CoViT latency without affecting its placement accuracy.

\subsubsection{Feature extractor}
The first part of the CoViT pipeline receives the assembled genome and outputs matrices that represent genome fragments. Feature extractor employs MinHash to find similar fragments (i.e., features) in different genomes.

% \begin{figure}[!t]
%   \begin{center}
%     \includegraphics[scale=0.2]{figs/FEE.png}
%   \end{center}
% \caption{Feature Extractor workflow example. $4-mers$ are extracted from the genome sequence using MinHash, where we apply $sketch$ and keep $n=3$ minimal hashed $4-mers$. Afterwards those $4-mers$ are used as anchors to generate the fragments (which are the 4-mers extended by 2 basepairs in each direction). The last step is one-hot encoding of the fragments.
% }
% \label{fig:fee}
% \end{figure}

\begin{algorithm}
\caption{the Feature Extractor}\label{alg:featureExtractor}
\begin{algorithmic}
\Require genome $g$, compression parameter $n$, fragment size $f$, k-mer size $k$, hash function $h$
\State $G \gets \{ g_i \equiv g[i: i+k]\} \text{ } s.t.\text{ } i\in [0, n-k+1]$
\State $kmers \gets sketch(G, n, h)$
\State $left \gets floor(\frac{f-k}{2})$
\State $right \gets ceil(\frac{f-k}{2})$
\State $fragments \gets \{g[i-left, i+right] : g_i \in kmers\}$\\
\Return $one-hot-encode(fragments)$
\end{algorithmic}
{\footnotesize One hot encoding of a fragment refers to the operation of transforming each base in the fragment as follows:\newline $A \rightarrow [1, 0, 0, 0]$; \textbf{  }
$C \rightarrow [0, 1, 0, 0]$; \textbf{  }
$G \rightarrow [0, 0, 1, 0]$; \textbf{  }
$T \rightarrow [0, 0, 0, 1]$; \textbf{  }
$N \rightarrow [0, 0, 0, 0]$.\newline For the example please refer to Figure~\ref{fig:ele}(a).}
\end{algorithm}

An example is illustrated in Figure~\ref{fig:fee}(b). Input is a genome sequence $g$ of length $N=19$. We generate all possible kmers $G$ (where $k=4$), and extract ($n=3$) 4-mers from the genome using the MinHash scheme (i.e., $sketch(G,n=3,h)$). The next step is to generate fragments of length $f=8$ (by extending each of three 4-mers by $\frac{f-k}{2}=\frac{8-4}{2}=2$ bases in each direction). The last step of this workflow is to encode each basepair of the fragments using one-hot encoding.

For the genome sequence $g$ of length $N$, the extractor first generates a set $G$ of all possible kmers (genome sub-sequences of length $k$). It then sketches (i.e., applies MinHash \emph{sketch} function to) the set of kmers, $G$, to extract $n$ representative kmers of the genome (Algorithm \ref{alg:sketch}). Those kmers are used as anchors to be expanded to generate fragments. Last part is transforming a fragment into a numerical matrix where the genome basepairs A, G, C and T are encoded using one-hot encoding as described in Algorithm~\ref{alg:featureExtractor}.

\subsubsection{Embedding layer}
\begin{figure}
  \begin{center}
    \includegraphics[scale=0.4]{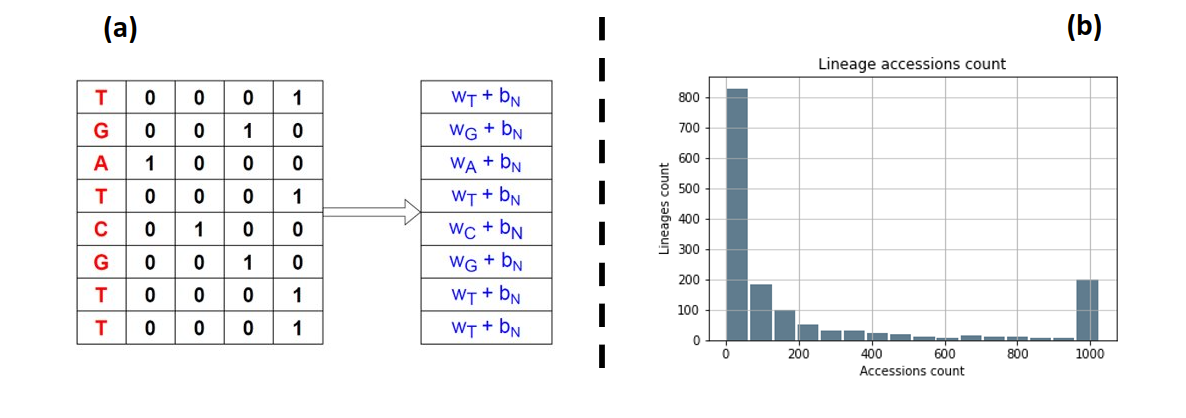}
  \end{center}
\caption{(a): Embedding a fragment matrix of dimensions $8\times4$. (b): The number of sequenced genomes per lineage histogram. 
Most of the lineages (more than 800 out of 1,536 available) have less than 64 accessions available. The peak in the last bin accounts for all lineages with more than 960 accessions.
}
\label{fig:ele}
\label{fig:hist}
\end{figure}

The input of the embedding layer is a genome fragment matrix of dimensions $f\times4$ (i.e., $f$ represents the fragment length and 4 is the dimensionality of the basepair one-hot encoding). It outputs a feature vector of size $f$. The embedding layer consists of one neuron which performs a base-wise linear transformation. Specifically, the embedding layer defines 5 learnable parameters, 4 weights $w_A, w_C, w_G, w_T$ and one bias term $b_N$. Given a genome fragment matrix, we transform each base (represented in one-hot encoding) to a numerical token as presented in Figure~\ref{fig:ele}(a), which  also contains an example in which we embed a fragment matrix of dimensions $8\times 4$.

% \begin{align*}
% \textbf{{\color{red}(A)}:  } [1, 0, 0, 0] & \rightarrow w_A + b_N\\
% \textbf{{\color{red}(C)}:  } [0, 1, 0, 0] & \rightarrow w_C + b_N\\
% \textbf{{\color{red}(G)}:  } [0, 0, 1, 0] & \rightarrow w_G + b_N\\
% \textbf{{\color{red}(T)}:  } [0, 0, 0, 1] & \rightarrow w_T + b_N\\
% \textbf{{\color{red}(N)}:  } [0, 0, 0, 0] & \rightarrow b_N.
% \end{align*}

\section{Evaluation}
  \subsection{Setup}
%https://www.overleaf.com/project/62984f7ba057d16cc916da1b
In this section we describe the dataset, evaluation criteria, training process and the machine on which we train the model.

%% TODO: Add the fact that you implemented CoViT using the TensorFlow framework in Python.
\subsubsection{Dataset}
We use the COVID-19 Data Portal~\cite{covid19data} as a source of sequenced SARS-CoV-2 genomes. There are 4,470,553 assembled genomes classified into 1,536 distinct lineages. The distribution of genomes across the lineages is highly non-uniform. For example, 463,426 samples are categorized as B.1.1.7, while the majority of the lineages has only a small number of samples, as depicted in Figure~\ref{fig:hist}(b). Such non-uniformity makes training the network challenging. To mitigate this, we select at most 1024 samples from each lineage and discard the rest when constructing the dataset. This way we prevent any potential over-fitting towards lineages like B.1.1.7. We also discard all samples that belong to lineages with fewer than 512 accessions, since such a small number of samples limits the model's ability to generalize. These steps form our dataset, which contains 240,734 samples belonging to 270 different lineage classes.

We divide our dataset into three non overlapping subsets: the training set, the validation set, and the test set. The training and validation sets are used to train and optimize the network. The test set is applied to evaluate CoViT's placement accuracy and to compare it with state-of-the-art solutions. Both the validation and test sets consist of 13,500 genomes randomly selected from the dataset. The rest of the samples comprise the training set. 

% \begin{figure}[hbt!]
%   \begin{center}
%     \includegraphics[scale=0.55]{figs/hist.png}
%   \end{center}
% \caption{Accession (the number of sequenced genomes per lineage) histogram. Most of the lineages (more than 800 out of 1,536 available) have less than 64 accessions available. The peak in the last bin accounts for all lineages with more than 960 accessions.
% }
% \label{fig:hist}
% \end{figure}

\subsubsection{Classification accuracy and placement rate criteria}
We use the \emph{top-n accuracy} as CoViT placement accuracy criterion. It estimates the likelihood of the \emph{correct} placement (i.e., the lineage to which the query genome belongs to\footnote{Since the COVID-19 Data Portal assigns all sequenced genomes to their respective lineages, the correct results are known.}) appearing among the \emph{n} most probable results (i.e., \emph{n} lineages output by the network at the top of the descending-order probability list). Specifically, we measure the top-1, top-2 and top-5 accuracy of our model as a criteria of its performance. Such criteria make the comparison to the state-of-the-art straightforward.

By \emph{base ambiguity} we mean the ambiguous bases (i.e. R, Y, K, N etc.) in assembled SARS-CoV-2 genomes. Base ambiguity is a result of incomplete coverage. N marks the worst ambiguity because it can mean any base (A, G, C, or T). We further evaluate the accuracy of CoViT placement under different levels of base ambiguity. In our study, we conservatively use only the worst case (i.e., N) ambiguity. 
% as well as different error rates.     

Another criteria we apply in our analysis is the \emph{placement rate}, a fraction of accessions a placement algorithm is able to place (correctly or incorrectly) on the existing phylogenetic tree. The reasons for the placement rate to be lower than 100\% are base ambiguity and sequencing errors.

\subsubsection{Network Training}
% Keep it short and simple!!!
We train CoViT using Adam~\cite{kingma2014adam} which is an optimized gradient descent technique, with $\beta_1 = 0.9, \beta_2=0.999$, batch size of 1024, and weight decay of 0.0001. Through various experiments, we found it most beneficial to choose the following hyper-parameters: $encoder repeats=4$, $d_{model}=256$, $d_{v} = 96$, $d_{k}= 96$, $d_{ff}=1536$, $h=18$. We also applied a dropout of rate 0.2 after each sub-layer.
The model was trained to minimize the categorical cross-entropy loss function. After fine-tuning and choosing those hyper-parameters, we train the model until the validation loss saturates.

\subsubsection{Hardware}
\label{HW}
We train CoViT on a desktop computer with Intel i7-9700K CPU with 8 cores, operating at 3.60GHz with 32GB of 2,666 MT/s DDR4 RAM, and NVIDIA's GeForce RTX 2080 Ti GPU with an 11GB / 14Gbps GDDR6 frame buffer, running at 1.545GHz.

\subsection{Results}

\subsubsection{Accuracy}
We measure CoViT performance using top-n accuracy.   
%as a placement tool, using the classification accuracy criterion. 
We compare CoViT to the state-of-the-art phylogenetic placement tool, UShER~\cite{turakhia2021ultrafast}, which uses a maximum parsimony approach where it searches for a placement that requires the fewest additional mutations. UShER achieves 92.1\% top-1 accuracy over the test set, while CoViT presents a 2.1\% increment over UShER in top-1 accuracy performance, achieving 94.2\%. Moreover, CoViT achieves a top-2 accuracy of 97.9\% and top-5 accuracy of 99.8\% as summarized in  Table~\ref{Tab:t1}.

\begin{table}[hbt!]
\caption{Accuracy comparison
\label{Tab:t1}}
\begin{center}
{
\begin{tabular}{||c|c|c|c||} 
 \hline
 Program & top-1 accuracy & top-2 accuracy & top-5 accuracy\\ [0.5ex] 
 \hline\hline
 CoViT & 94.2\% & 97.9\% & 99.8\%\\ [1ex]
 \hline
 UShER & 92.1\% & - & -\\ [1ex]
 \hline
\end{tabular}
}
\end{center}
{\footnotesize 
%A less harsh performance criterion than the top-1 accuracy is used by UShER. In which, rather than measuring the amount of correct lineage classifications, they measure the amount of correct sister node classifications. 
UShER uses another accuracy criterion called \emph{sister node placement}. They report the correct sister node placement in 97.2\% of cases. While this criterion resembles the top-2 accuracy, they are not identical, therefore we do not report it as a top-2 accuracy for UShER.}

\end{table}

\subsubsection{Run time and memory usage}
\label{efficiency}
We measure CoViT run time and memory usage and compare it to SARS-CoV-2 state-of-the-art placement tools. We compare CoViT and UShER, classifying the genomes in the test set, running on the machine specified in~\ref{HW} (see Table~\ref{Tab:t2}). We also present in Table~\ref{Tab:t2} the comparison to IQ-TREE multicore~\cite{nguyen2015iq}, and EPA-ng~\cite{barbera2019epa} running on a server with 160 processors (Intel Xeon CPU E7-8870 v.4, 2.10 gigahertz), each with 20 CPU cores. Those results where obtained while adding just one sequence to the SARS-CoV-2 global phylogeny containing 38,342 leaves~\cite{turakhia2021ultrafast}.

% \begin{table}[hbt!]
% \caption{Run-time and memory requirements comparison
% \label{Tab:t2}}
% \begin{center}
% \begin{tabular}{|p{4.3cm}|||p{1.3cm}|p{1.3cm}|p{1.3cm}||p{1.7cm}|p{1.7cm}|} 
%  \hline
%  \multicolumn{1}{|c|||}{Machine $\longrightarrow$} & 
% \multicolumn{3}{c||}{As specified in section~\ref{HW}} &
% \multicolumn{2}{c|}{ Adopted from~\cite{turakhia2021ultrafast}}\\
% \hline
%  \centering Program $\longrightarrow$ & CoViT (GPU) & CoViT (CPU) & UShER v0.5.6 & IQ-TREE multicore v2.1.1 & EPA-ng v0.3.8\\ [0.5ex]
%  \hline
%  Avg run-time per sample & 0.055s & 0.094s & 0.195s & 2791s & 1658s\\
%  \hline
%  Peak memory used (GB) & 5.02 & 3.74 & 0.49 & 12.85 & 791.82\\
%  \hline
%  Speedup over UShER & 3.53 & 2.06 & 1 & $7\cdot10^{-6}$ & $6\cdot10^{-5}$\\
%  \hline
% \end{tabular}
% \end{center}
% \end{table}

\newcommand{\mc}[2]{\multicolumn{#1}{c}{#2}}
\definecolor{Gray}{gray}{0.85}
\definecolor{LightCyan}{rgb}{0.88,1,1}

\begin{table}[hbt!]
\caption{Run-time and memory requirements comparison}
\label{Tab:t2}
\begin{center}
{
\begin{tabular}{|c|c|c|c|}
\hline
 Program & Avg run-time per sample & Peak memory used (GB) & speedup over UShER\\ [0.5ex]
 \hline
 \hline
 \rowcolor{LightCyan}CoViT (GPU) & \textbf{0.055s} & 5.02 & \textbf{3.53}\\
 \hline
 \rowcolor{LightCyan}CoViT (CPU) & 0.094s & 3.74 & 2.06\\
 \hline
 \rowcolor{LightCyan}UShER v0.5.6 & 0.195s & \textbf{0.49} & 1\\
 \hline
 \hline
 \rowcolor{yellow}IQ-TREE multicore v2.1.1 & 2791s & 12.85 & $7\cdot10^{-6}$\\
 \hline
  \rowcolor{yellow}EPA-ng v0.3.8 & 1658s & 791.82 & $6\cdot10^{-5}$\\
 \hline
%  Speedup over UShER & 3.53 & 2.06 & 1 & $7\cdot10^{-6}$ & $6\cdot10^{-5}$\\
%  \hline
\end{tabular}
}
\end{center}
{\footnotesize The results marked in light cyan are given running of the machine specified in section~\ref{HW}. The results marked in yellow are adopted from~\cite{turakhia2021ultrafast}.}
\end{table}

\subsubsection{Robustness to base ambiguity}
\begin{figure}[hbt!]
  \begin{center}
    \includegraphics[scale=0.32]{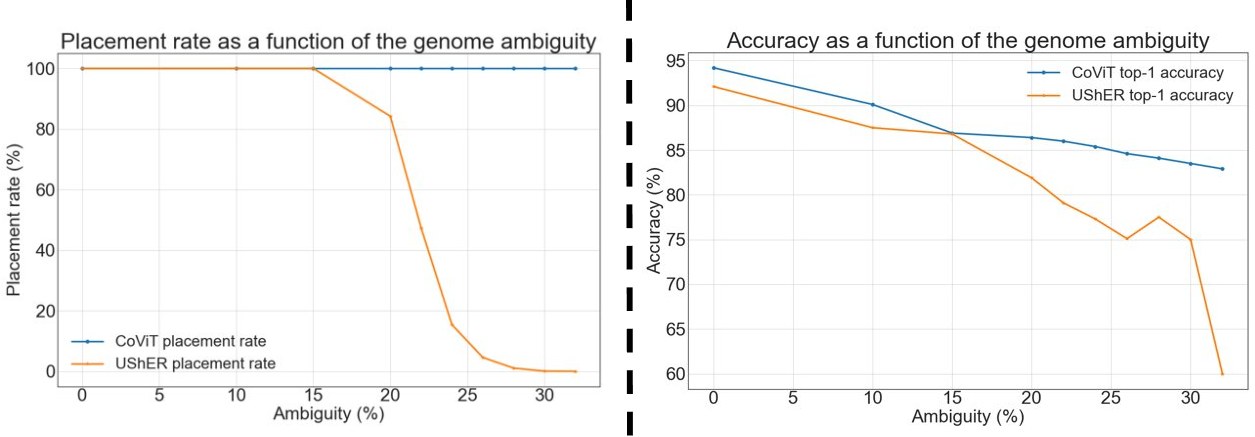}
  \end{center}
\caption{Placement rate (left) and accuracy (right) as a function of the genome ambiguity. i.e., the fraction of the ambiguous bases in the input genome.
}
\label{fig:ambig}
\end{figure}

Some low quality SARS-CoV-2 genomes  may contain ambiguous bases, which is a result of incompletely covered genome or sequencing errors. It is highly desirable that such genomes are correctly placed even when the level of ambiguity is significant. To test the robustness of CoViT to base ambiguity, we mask  random sites (i.e., replace basepairs by N) in each SARS-CoV-2 genome of the test set, with the percentage of ambiguous bases ranging from 0 to 32\%. We then compare CoViT accuracy and placement rate on this \emph{ambiguous} test set (13,500 accessions with randomly masked sites) against the placement rate and accuracy of UShER. Figure~\ref{fig:ambig} presents the results.
At 32\% ambiguity, UShER's placement rate is 0.5\%, which means UShER places only 67 accessions out of 13,500. For the rest of accessions, UShER returns "failed to map" message.  For those accessions UShER is able to map, it reaches the top-1 accuracy of 60.5\%.
CoViT maintains a constant 100\% placement rate. At 32\% ambiguity, CoViT obtains the top-1 accuracy of 82.9\%. 
CoViT maintains 100\% placement rate up to 50\% ambiguity (not shown in the figure), at which point CoViT's top-1 accuracy stands at 77.3\%. CoViT exhibits such resilience to base ambiguity because neural networks typically maintain high resilience when processing low quality (high error or ambiguity rates) data. 

% \subsubsection{Sensitivity to high error rates}

\section{Conclusion}
%We presented CoViT, a software tool that places viral genomes such as SARS-CoV-2 on existing phylogenetic trees. Unlike current solutions, CoViT is a deep neural network based algorithm. 
The massive amount of sequenced SARS-CoV-2 DNA material paves the way for real-time surveillance of the virus evolution and spread. At the same time, it puts a tremendous strain on phylogenetic analysis infrastructures. To prevent losing valuable genomic data and reaching incorrect conclusions, a new computing paradigm for phylogenetic analysis of viral pandemics is highly required. 
CoViT is inspired by an enormous success of neural networks in implementing classification tasks. 
%CoViT applies methods of image classification to phylogenetic analysis of viral genome and thus provides the much sought after ability to incorporate viral accessions on a global phylogenetic tree quickly and accurately. 
CoViT extracts sets of features from a SARS-CoV-2 accession genome and uses Vision Transformer, an attention based neural network model, to accurately classify and place it in real-time. CoViT not only outperforms state-of-the-art tools both in accuracy and run-time, by placing a SARS-CoV-2 genome on a phylogenetic tree in 0.055s with 94.2\% accuracy. It also shows a strong resilience to low quality genome data (i.e. high ambiguity of the newly sequenced accessions). CoViT is able to place SARS-CoV-2 genomes with 50\% ambiguity (when N replaces A, G, C, and T in 50\% of the sites in each genome), while maintaining 77.3\% accuracy.     

CoViT lifts the computational barrier to pandemic surveillance by improving the placement accuracy. It exhibits  highly robust performance under low quality and ambiguous genomic data while greatly decreasing the processing time. Thus it enables real-time evolutionary tracking of COVID-19 and future viral pandemics.

% \section{Evaluation - old version}
%   \input{results_old.tex}
% \section{Future work}
% \input{future.tex}

% \section*{Acknowledgements}

\bibliographystyle{myrecomb}
\bibliography{mybib}

\end{document}